\documentclass[runningheads]{llncs} 
\usepackage[T1]{fontenc}

\usepackage{graphicx}
\usepackage[hidelinks,colorlinks=true,citecolor=blue, linkcolor=blue, urlcolor=blue]{hyperref}
\usepackage{color}

\usepackage[numbers]{natbib}
\setcitestyle{square}

\usepackage{subcaption}
\usepackage{amsmath}

\usepackage[capitalize]{cleveref} 
\crefname{section}{Sec.}{Secs.}
\Crefname{section}{Section}{Sections}
\Crefname{table}{Table}{Tables}
\crefname{table}{Tab.}{Tabs.}

\usepackage{tabularx}
\usepackage[]{acronym}

\acrodef{atc}[ATC]{Automatic Thresholding Criterion}
\acrodef{bsds500}[BSDS500]{Berkeley Segmentation Data Set}
\acrodef{csa}[CSA]{Cuckoo Search Algorithm}
\acrodef{fsim}[FSIM]{Feature Similarity Index}
\acrodef{mec}[MEC]{Maximum Entropy Criterion}
\acrodef{met}[MET]{Minimum Error Thresholding}
\acrodef{psnr}[PSNR]{Peak Signal-to-Noise Ratio}
\acrodef{ssim}[SSIM]{Structural Similarity Index}

\begin{document}
\title{Image Thresholding: Understanding Bias of Evaluation Metrics towards Specific Evaluation Functions}
\titlerunning{Understanding Bias of Evaluation Metrics towards Evaluation Functions}

\author{
    Eslam Hegazy\inst{1}\orcidID{0009-0007-5084-4702} \and
    Mohamed Gabr\inst{1}\orcidID{0000-0003-3690-9585}
}

\authorrunning{E. Hegazy and M. Gabr}

\institute{German University in Cairo, Cairo, Egypt
\email{\{eslam.sabry,mohamed.kamel-gabr\}@guc.edu.eg}\\
}
\maketitle
\begin{abstract}
Multilevel image thresholding is widely used for segmentation in applications ranging from medical imaging to remote sensing. Classical objective functions, such as Otsu's between-class variance and Kapur's entropy, are often optimized using metaheuristic algorithms, with performance evaluated via metrics like Structural Similarity Index (SSIM) and Peak Signal-to-Noise Ratio (PSNR). These evaluations implicitly assume that SSIM and PSNR provide unbiased measures of segmentation quality.
In this study, we examine this assumption by analyzing the correlation between thresholding objective functions and quality metrics across all possible thresholds for images in the BSDS500 dataset. Results show that Otsu's criterion consistently exhibits high correlation with both SSIM and PSNR, while Kapur's entropy demonstrates weaker and more variable correlation. Otsu outperforms Kapur in correlation with PSNR for all images and with SSIM for over 91\%.
Our findings reveal an inherent metric–objective-function bias. This work highlights the need for more neutral evaluation frameworks and motivates extending the analysis to additional thresholding criteria and domains. Source code of this paper can be found at \url{https://w3id.org/met-dp/icpr26-95}.

\end{abstract}
\section{Introduction}

Image thresholding remains one of the most fundamental techniques in image segmentation, owing to its simplicity, computational efficiency, and wide applicability across domains such as medical imaging, remote sensing, document analysis, and industrial inspection \cite{med,sat,defect}. Thresholding methods can be broadly categorized into binary thresholding, where a single threshold partitions the image into foreground and background, and multilevel thresholding, which generalizes this concept by determining multiple thresholds to segment an image into several meaningful regions.

Historically, multilevel thresholding approaches have been developed as extensions of binary thresholding formulations. While binary thresholding methods can often be solved optimally through exhaustive search, the computational complexity of multilevel thresholding grows exponentially with the number of thresholds \citep{Tuba_2014}. This combinatorial explosion renders exact optimization impractical for real-world applications, motivating the adoption of optimization-based approaches.

In recent decades, a large body of literature has focused on employing metaheuristic optimization algorithms, such as genetic algorithms, particle swarm optimization, differential evolution, firefly algorithm, whale optimization, and their numerous variants, to efficiently approximate optimal threshold values. These algorithms typically optimize a predefined thresholding evaluation (objective) function, most notably Otsu's between-class variance \citep{otsu} and Kapur's entropy-based criterion \citep{kapur}. The primary motivation behind these approaches is to trade strict optimality for significant gains in computational efficiency and scalability \citep{mealpy}.

A common research pattern has emerged in this domain: proposing a new optimization algorithm, a modified version of an existing algorithm, or a tuned set of hyperparameters, and applying it to optimize a known (or slightly modified) thresholding evaluation function. The resulting method is then evaluated on a set of benchmark images and compared against other algorithm–objective-function combinations. Performance is typically assessed using two main criteria: thresholding quality and computational runtime.

Thresholding quality is most commonly quantified using full-reference image quality metrics, particularly \acf{ssim} and \acf{psnr} \citep{Amiriebrahimabadi_Rouhi_Mansouri_2024}. Due to the stochastic nature of metaheuristic algorithms, statistical significance tests are often conducted to support the robustness of the reported results.

Despite the extensive volume of comparative studies, an implicit assumption is frequently made: that the employed quality metrics (e.g., \ac{ssim} and \ac{psnr}) provide an objective and neutral basis for comparing different thresholding strategies and objective functions. However, empirical evidence across the literature suggests a recurring trend: thresholding methods based on Otsu's criterion often achieve higher \ac{ssim} and \ac{psnr} values than those based on Kapur's entropy, while in certain scenarios the opposite behavior is observed. These inconsistencies are typically attributed to image characteristics or optimization performance, rather than to the evaluation metrics themselves.

For instance, \citet{Kalyani_2021} present an advanced approach to color image segmentation by utilizing multilevel thresholding enhanced by the \ac{csa}. To identify optimal image divisions, the study evaluates three distinct mathematical frameworks: Kapur's entropy, Otsu's between-class variance, and minimum cross entropy. The authors demonstrate that the \ac{csa} effectively navigates complex search spaces using Levy flights, which prevents the system from becoming stuck in suboptimal data points. Experimental results across various standard test images reveal that the Otsu-based method consistently delivers the most accurate and reliable segmentation performance. Furthermore, the study utilizes quality metrics like \ac{ssim} and \ac{psnr} to confirm that this optimized technique is highly effective for high-precision tasks such as medical imaging and computer vision. 
\citet{samantaray2020new} introduce a hybrid optimizer called Harris Hawks Cuckoo Search (HHO-CS) for multilevel thresholding of RGB thermogram images, using both 1D Otsu's between-class variance and Kapur's entropy as fitness (objective) functions to compute optimal thresholds and compare performance fairly across criteria. The authors evaluate the proposed HHO-CS against the original Harris Hawks Optimization (HHO) and the Cuckoo Search (CS) algorithms on thermogram segmentation tasks using each of these objective functions. Results show HHO-CS consistently outperformed HHO and CS under each of Otsu's and Kapur's criteria, and the segmentation quality achieved using Otsu's between-class variance is generally better than using Kapur's entropy for these infrared images. The method was evaluated with \ac{psnr}, \ac{ssim}, and \ac{fsim} metrics to confirm higher image quality after segmentation.

This observation motivates the central hypothesis of the present study: The relative performance of thresholding objective functions, as measured by \ac{ssim} and/or \ac{psnr}, is inherently scenario-dependent and systematically biased toward certain objective functions.
More specifically, we hypothesize that for a given testing scenario (or set of image characteristics), \ac{ssim} and \ac{psnr} consistently favor one thresholding objective function (e.g., Otsu) over others (e.g., Kapur), regardless of the optimization algorithm used. Consequently, when an optimization algorithm is applied to a thresholding objective function that is already favored by the evaluation metric, the resulting method will naturally appear superior in comparative studies. Conversely, optimization over an objective function disfavored by the metric may be unjustly perceived as inferior.
Under this perspective, observed performance gains attributed to a novel optimization algorithm may, in part, be explained by an underlying metric–objective-function bias, rather than by genuine improvements in threshold selection. This raises an important question regarding the fairness and interpretability of comparative studies in multilevel image thresholding:
Are \ac{ssim} and \ac{psnr} unbiased indicators of thresholding quality across different objective functions, or do they inherently favor specific thresholding functions?

The contribution of this work lies in explicitly investigating this hidden factor by decoupling the effects of the optimization algorithm from those of the thresholding objective function and the evaluation metric. By systematically analyzing the relationship between objective functions and quality metrics across multiple scenarios, this study aims to provide deeper insight into the validity of commonly used evaluation practices and to highlight potential biases that may influence reported conclusions in the literature.

\section{Background}

\subsection{\acf{ssim}}

SSIM is generally defined by the global equation in \cref{eq:ssim_global} \citep{wang2004}:
\begin{equation}
\label{eq:ssim_global}
\mathrm{SSIM}(x, y) =
\frac{(2\mu_x \mu_y + C_1)(2\sigma_{xy} + C_2)}
{(\mu_x^2 + \mu_y^2 + C_1)(\sigma_x^2 + \sigma_y^2 + C_2)},
\end{equation}
where \( \mu_x \) and \( \mu_y \) denote the mean intensities of images \( x \) and \( y \), respectively, \( \sigma_x^2 \) and \( \sigma_y^2 \) their variances, and \( \sigma_{xy} \) their covariance. The constants \( C_1 \) and \( C_2 \) are included to stabilize the division in regions of low luminance or contrast and are defined as \( C_1 = (K_1 L)^2 \) and \( C_2 = (K_2 L)^2 \), where \( L \) is the dynamic range of the image and \( K_1 \) and \( K_2 \) are small positive constants.

However, from a practical perspective and as emphasized in the original work of \citet{wang2004}, SSIM is not evaluated using global image statistics. Instead, it is computed locally over a sliding window and subsequently averaged across the spatial domain. This local formulation is also adopted in the experiments of this paper, following the implementation provided by the \texttt{scikit-image} Python library.

Let \( p \) denote a pixel location and \( \mathcal{N}_p \) a local neighborhood centered at \( p \), defined by a window of fixed size. The default size in \texttt{scikit-image} library is 7. This size is also the one used throughout experiments of this paper. The local means are computed as
\begin{equation}
\mu_x(p) = \sum_{q \in \mathcal{N}_p} w(q)\,x(q),
\quad
\mu_y(p) = \sum_{q \in \mathcal{N}_p} w(q)\,y(q),
\end{equation}
where \( w \) denotes a normalized weighting function satisfying \( \sum_q w(q) = 1 \). In practice, \( w \) is either a uniform or a Gaussian window.

The corresponding local variances and covariance are given by
\begin{equation}
\sigma_x^2(p) = \sum_{q \in \mathcal{N}_p} w(q)\,[x(q) - \mu_x(p)]^2,
\end{equation}
\begin{equation}
\sigma_y^2(p) = \sum_{q \in \mathcal{N}_p} w(q)\,[y(q) - \mu_y(p)]^2,
\end{equation}
\begin{equation}
\sigma_{xy}(p) = \sum_{q \in \mathcal{N}_p} w(q)\,[x(q) - \mu_x(p)][y(q) - \mu_y(p)].
\end{equation}

Using these local statistics, the SSIM map is defined at each pixel location as
\begin{equation}
\label{eq:ssim_local}
\mathrm{SSIM}(p) =
\frac{
\left(2\mu_x(p)\mu_y(p) + C_1\right)
\left(2\sigma_{xy}(p) + C_2\right)
}{
\left(\mu_x^2(p) + \mu_y^2(p) + C_1\right)
\left(\sigma_x^2(p) + \sigma_y^2(p) + C_2\right)
}.
\end{equation}

Finally, the overall SSIM score is obtained by averaging the local SSIM values over all pixel locations:
\begin{equation}
\mathrm{SSIM}(x, y) =
\frac{1}{|\Omega|}
\sum_{p \in \Omega} \mathrm{SSIM}(p),
\end{equation}
where \( \Omega \) denotes the set of all spatial locations in the image. 

\subsection{\acf{psnr}}
The \ac{psnr} is a commonly used full-reference image quality metric that quantifies the fidelity between a reference image \( x \) and a distorted or reconstructed image \( y \). It is defined in terms of the mean squared error (MSE) as
\begin{equation}
\label{eq:psnr}
\mathrm{PSNR}(x, y) = 10 \log_{10} \left( \frac{L^2}{\mathrm{MSE}(x, y)} \right),
\end{equation}
where \( L \) denotes the maximum possible pixel intensity value of the image.

The mean squared error is given by
\begin{equation}
\label{eq:mse}
\mathrm{MSE}(x, y) = \frac{1}{HW}
\sum_{i=1}^{H} \sum_{j=1}^{W}
\left( x_{i,j} - y_{i,j} \right)^2,
\end{equation}
with \( H \) and \( W \) representing the height and width of the image, respectively. For multichannel images, the MSE is typically computed independently for each channel and averaged across channels.

The \ac{psnr} expresses the ratio between the peak signal power, represented by \( L^2 \), and the distortion power measured by the MSE on a logarithmic decibel scale. Higher \ac{psnr} values indicate lower reconstruction error and thus higher image fidelity. Despite its simplicity and widespread use, the \ac{psnr} correlates imperfectly with human visual perception, as it penalizes all pixel-wise deviations equally and does not explicitly model structural or perceptual characteristics of images.

\subsection{Otsu Thresholding Function}
Otsu's method is a non-parametric and unsupervised thresholding technique that determines optimal threshold values by maximizing the separability between classes in the image intensity histogram. Originally proposed for bi-level thresholding, the method assumes that the image can be partitioned into classes corresponding to different intensity ranges and selects thresholds that maximize the between-class variance, which is equivalent to minimizing the within-class variance. Due to its simplicity and effectiveness, Otsu's method has been widely adopted for image segmentation and preprocessing tasks.

Let an image have \( L \) discrete gray levels \( \{0, 1, \ldots, L-1\} \), and let \( p(i) \) denote the normalized histogram, such that \( \sum_{i=0}^{L-1} p(i) = 1 \). For multilevel thresholding with \( K \) thresholds \( \{t_1, t_2, \ldots, t_K\} \), the image is partitioned into \( K+1 \) classes. The probability (weight) and mean intensity of class \( k \) are defined as
\begin{equation}
\omega_k = \sum_{i=t_{k-1}}^{t_k - 1} p(i),
\qquad
\mu_k = \frac{1}{\omega_k} \sum_{i=t_{k-1}}^{t_k - 1} i\, p(i),
\end{equation}
where \( t_0 = 0 \) and \( t_{K+1} = L \). The global mean intensity is given by
\begin{equation}
\mu_T = \sum_{i=0}^{L-1} i\, p(i).
\end{equation}

The optimal multilevel thresholds are obtained by maximizing the between-class variance, defined as
\begin{equation}
\label{eq:otsu_multilevel}
\sigma_B^2 = \sum_{k=1}^{K+1} \omega_k (\mu_k - \mu_T)^2.
\end{equation}
The set of thresholds \( \{t_1^\ast, \ldots, t_K^\ast\} \) that maximizes \( \sigma_B^2 \) is selected as the solution. While the exhaustive search required for multilevel Otsu thresholding grows exponentially with the number of thresholds, various optimization strategies have been proposed to reduce computational complexity, enabling its application in practical multilevel segmentation scenarios.

\subsection{Kapur Thresholding Function}
Kapur's method is an unsupervised, histogram-based thresholding technique that selects optimal threshold values by maximizing the entropy of segmented classes. Unlike variance-based approaches such as Otsu's method, Kapur's criterion is grounded in information theory and seeks thresholds that maximize the information content of the resulting partition. Owing to its robustness and conceptual simplicity, Kapur's method has been widely used for image segmentation and multilevel thresholding tasks.

Let an image have \( L \) discrete gray levels \( \{0, 1, \ldots, L-1\} \), and let \( p(i) \) denote the normalized histogram, such that \( \sum_{i=0}^{L-1} p(i) = 1 \). For multilevel thresholding with \( K \) thresholds \( \{t_1, t_2, \ldots, t_K\} \), the image is partitioned into \( K+1 \) disjoint classes. The probability (weight) of class \( k \) is defined as
\begin{equation}
\omega_k = \sum_{i=t_{k-1}}^{t_k - 1} p(i),
\end{equation}
where \( t_0 = 0 \) and \( t_{K+1} = L \). The entropy of class \( k \) is given by
\begin{equation}
H_k = - \sum_{i=t_{k-1}}^{t_k - 1}
\frac{p(i)}{\omega_k}
\log \left( \frac{p(i)}{\omega_k} \right).
\end{equation}

The optimal multilevel thresholds are obtained by maximizing the sum of class entropies,
\begin{equation}
\label{eq:kapur_multilevel}
H = \sum_{k=1}^{K+1} H_k,
\end{equation}
and the set of thresholds \( \{t_1^\ast, \ldots, t_K^\ast\} \) that maximizes \( H \) is selected as the solution. Similar to multilevel Otsu thresholding, the computational complexity of Kapur's method increases exponentially with the number of thresholds, motivating the use of heuristic and metaheuristic optimization techniques in practical applications.

\subsection{Pearson Correlation Coefficient}

The linear correlation between two signals \( x \) and \( y \) is quantified using the Pearson correlation coefficient, which measures the strength of their linear dependence. It is defined as
\begin{equation}
\label{eq:pearson_corr}
\rho(x, y) =
\frac{
\sum_{i=1}^{N} (x_i - \mu_x)(y_i - \mu_y)
}{
\sqrt{
\sum_{i=1}^{N} (x_i - \mu_x)^2
\sum_{i=1}^{N} (y_i - \mu_y)^2
}
},
\end{equation}
where \( \mu_x \) and \( \mu_y \) denote the sample means of \( x \) and \( y \), respectively, and \( N \) is the number of samples. The coefficient \( \rho \in [-1, 1] \), with values close to \( 1 \) or \( -1 \) indicating strong positive or negative linear correlation, respectively, and values near zero indicating weak linear dependence.

\section{Experiment and Results}

We perform an experiment to evaluate the correlation between each of the evaluation functions of Otsu and Kapur and each of the \ac{ssim} and \ac{psnr} metrics. We use the \ac{bsds500} dataset \citep{bsds500}, consisting of 500 natural images. For each image, we test each possible threshold T where $0 < T \leq L-1$. For each threshold, we compute the value of Otsu, Kapur, \ac{ssim} and \ac{psnr}. Then, we compute the Pearson correlation coefficients of $\{\textsc{Otsu}, \textsc{Kapur}\} \times \{\textsc{SSIM}, \textsc{PSNR} \}$ over all possible thresholds in the same image. Lastly, we study outputs from all images of the dataset.

Now, the rest of this section is structured as follows. In \cref{subsec:sample_outputs}, we show representative sample outputs to illustrate typical behaviors observed in the data. Next, we analyze the global distribution of results to identify overall trends through \cref{subsec:histogram-cnts}. Finally, we summarize the key quantitative outcomes in tabular form, highlighting the main results of this research in \cref{subsec:quantit-tables}.

\subsection{Sample Outputs}
\label{subsec:sample_outputs}

\begin{figure}
    \centering
    \begin{subfigure}[b]{0.47\textwidth}
        \centering
        \includegraphics[height=0.5\textwidth,width=\textwidth]{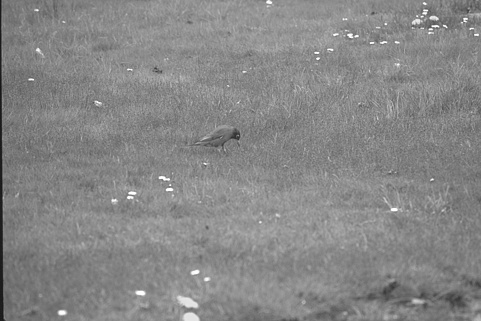}
        \caption{Image}
    \end{subfigure}
    \hfill
    \begin{subfigure}[b]{0.47\textwidth}
        \centering
        \includegraphics[width=\textwidth]{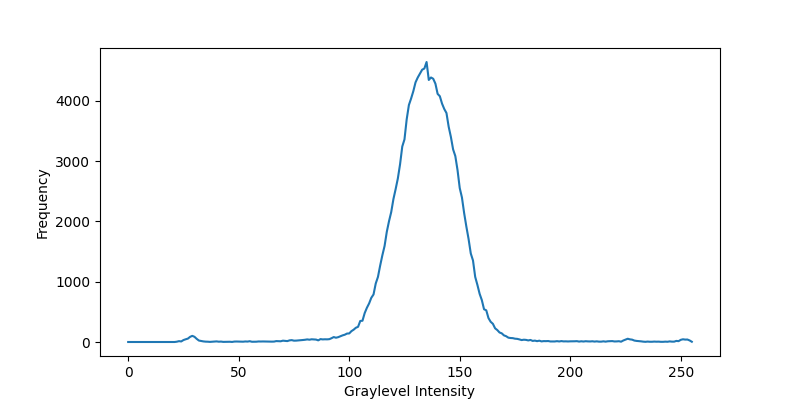}
        \caption{Histogram}
    \end{subfigure}
    \begin{subfigure}[b]{0.47\textwidth}
        \centering
        \includegraphics[width=\textwidth]{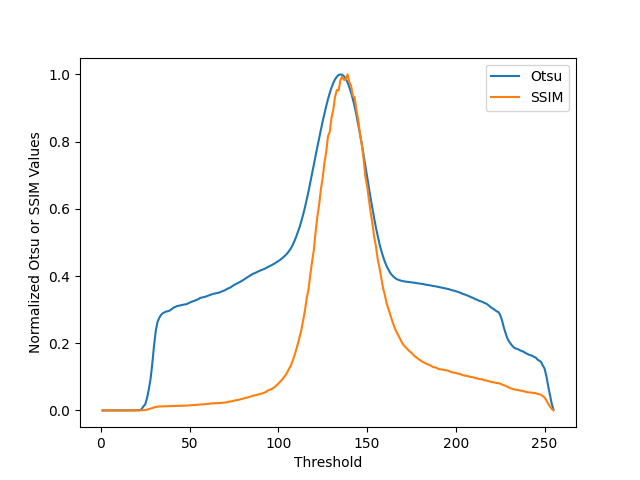}
        \caption{Otsu $\times$ SSIM. Correlation = 0.89.}
    \end{subfigure}
    \hfill
    \begin{subfigure}[b]{0.47\textwidth}
        \centering
        \includegraphics[width=\textwidth]{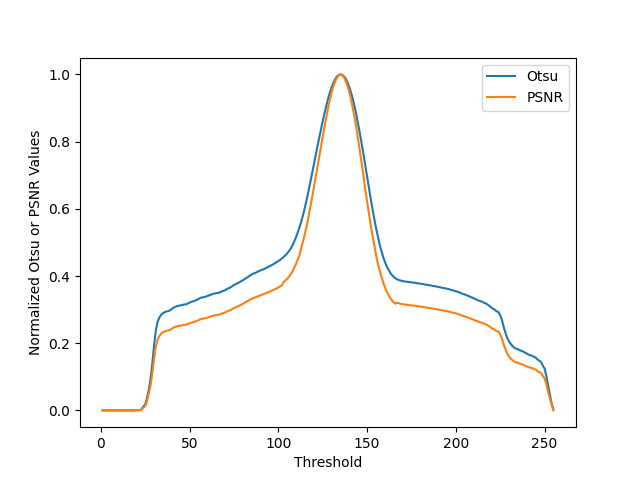}
        \caption{Otsu $\times$ PSNR. Correlation = 0.99.}
    \end{subfigure}


    \begin{subfigure}[b]{0.47\textwidth}
        \centering
        \includegraphics[width=\textwidth]{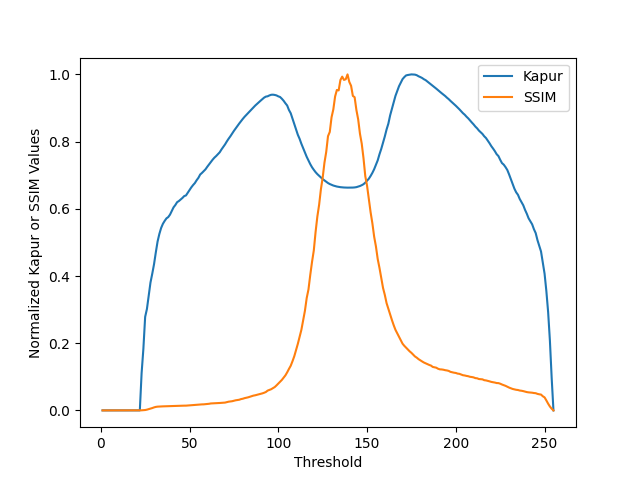}
        \caption{Kapur $\times$ SSIM. Correlation = 0.14.}
    \end{subfigure}
    \hfill
    \begin{subfigure}[b]{0.47\textwidth}
        \centering
        \includegraphics[width=\textwidth]{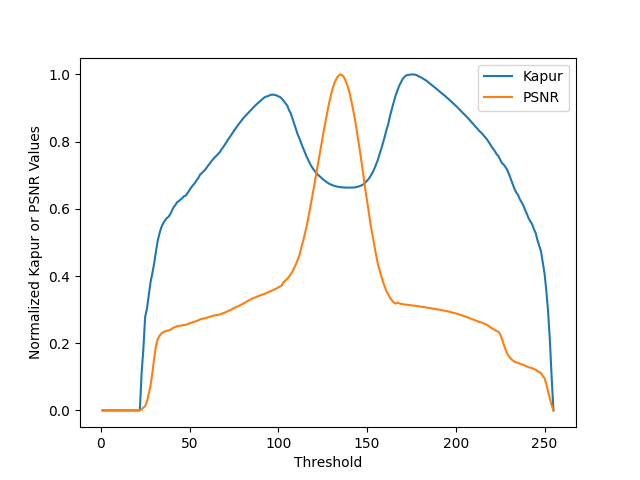}
        \caption{Kapur $\times$ PSNR. Correlation = 0.42.}
    \end{subfigure}
    \caption{Values of Otsu, Kapur, \ac{ssim} and \ac{psnr} on image 28096, from \ac{bsds500}, plotted versus each other. The values of each function/metric are normalized between 0 and 1 for clarity of visualization.}
    \label{fig:indiv_28096}
\end{figure}

\begin{figure}
    \centering
    \begin{subfigure}[b]{0.47\textwidth}
        \centering
        \includegraphics[height=0.5\textwidth,width=\textwidth]{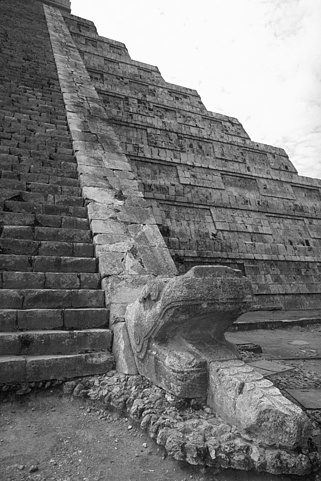}
        \caption{Image}
    \end{subfigure}
    \hfill
    \begin{subfigure}[b]{0.47\textwidth}
        \centering
        \includegraphics[width=\textwidth]{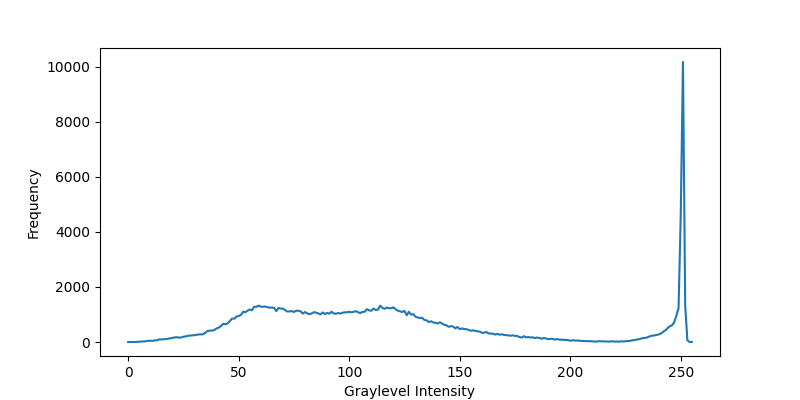}
        \caption{Histogram}
    \end{subfigure}
    \begin{subfigure}[b]{0.47\textwidth}
        \centering
        \includegraphics[width=\textwidth]{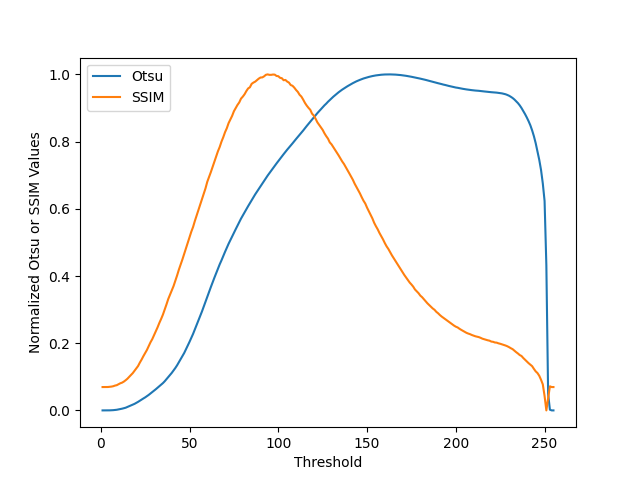}
        \caption{Otsu $\times$ SSIM. Correlation = 0.23.}
    \end{subfigure}
    \hfill
    \begin{subfigure}[b]{0.47\textwidth}
        \centering
        \includegraphics[width=\textwidth]{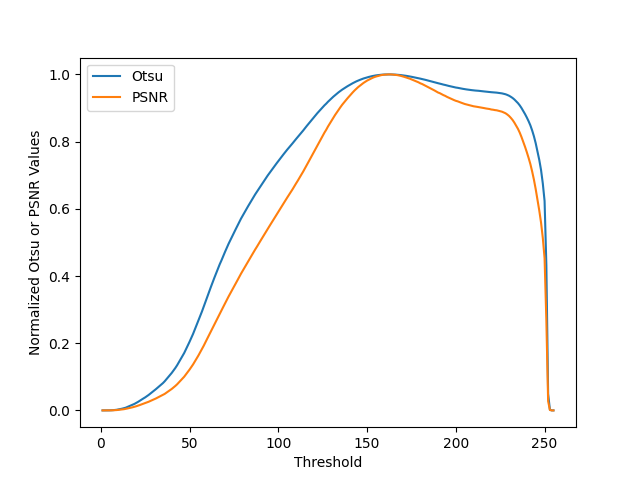}
        \caption{Otsu $\times$ PSNR. Correlation = 0.99.}
    \end{subfigure}


    \begin{subfigure}[b]{0.47\textwidth}
        \centering
        \includegraphics[width=\textwidth]{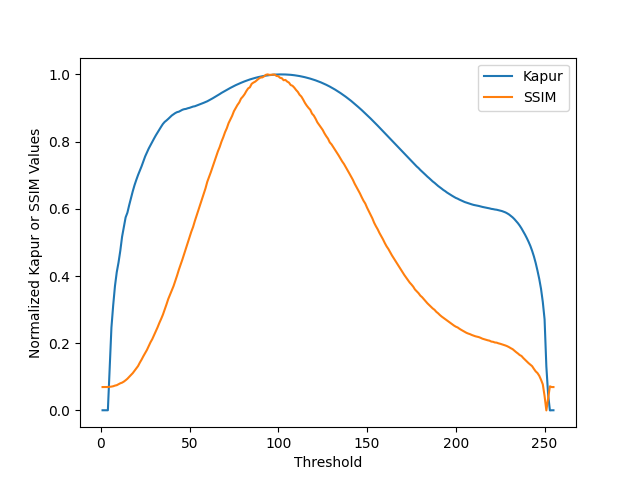}
        \caption{Kapur $\times$ SSIM. Correlation = 0.84.}
    \end{subfigure}
    \hfill
    \begin{subfigure}[b]{0.47\textwidth}
        \centering
        \includegraphics[width=\textwidth]{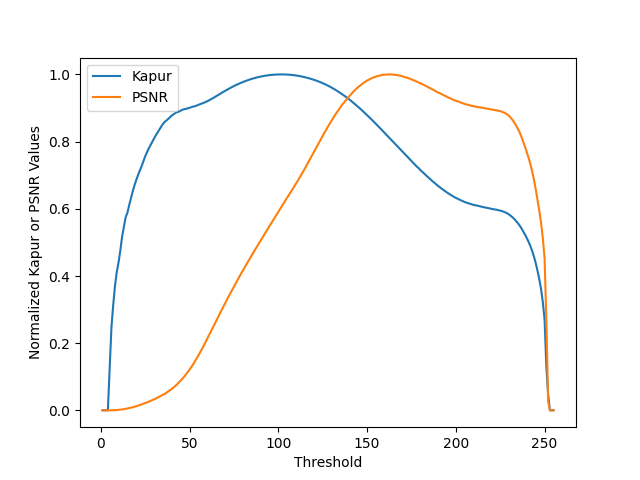}
        \caption{Kapur $\times$ PSNR. Correlation = 0.13.}
    \end{subfigure}
    \caption{Values of Otsu, Kapur, \ac{ssim} and \ac{psnr} on image 33044, from \ac{bsds500}, plotted versus each other. The values of each function/metric are normalized between 0 and 1 for clarity of visualization.}
    \label{fig:indiv_33044}
\end{figure}

\begin{figure}
    \centering
        \begin{subfigure}[b]{0.47\textwidth}
        \centering
        \includegraphics[height=0.5\textwidth,width=\textwidth]{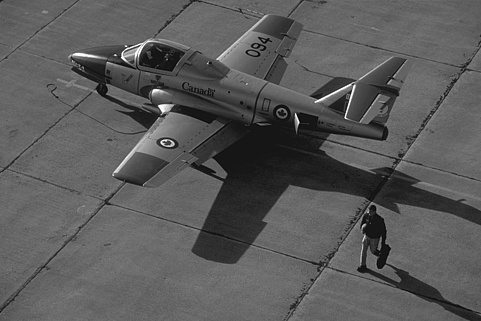}
        \caption{Image}
    \end{subfigure}
    \hfill
    \begin{subfigure}[b]{0.47\textwidth}
        \centering
        \includegraphics[width=\textwidth]{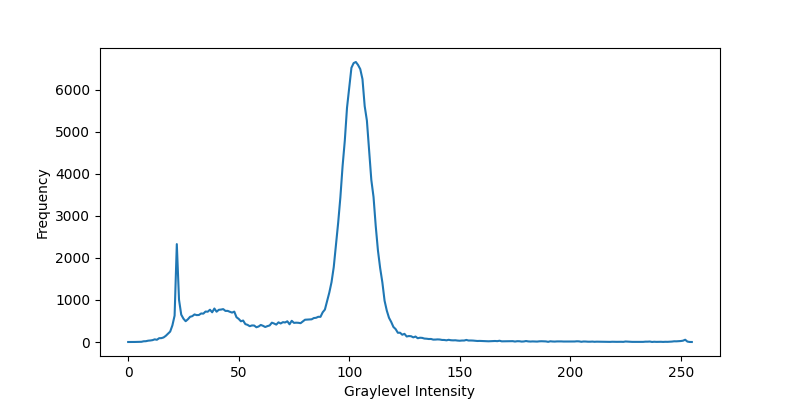}
        \caption{Histogram}
    \end{subfigure}

    \begin{subfigure}[b]{0.47\textwidth}
        \centering
        \includegraphics[width=\textwidth]{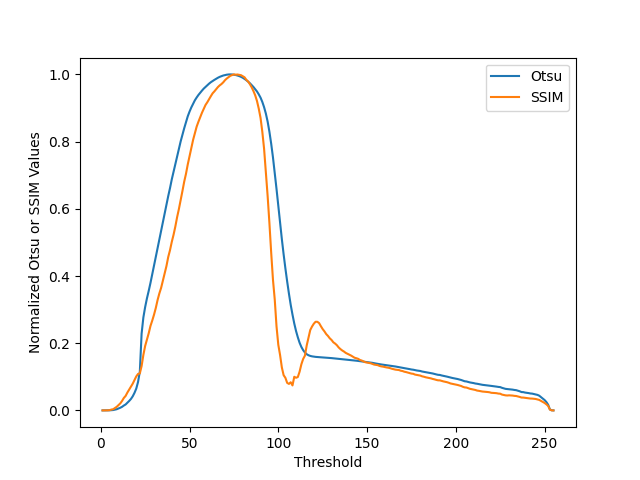}
        \caption{Otsu $\times$ SSIM. Correlation = 0.97.}
    \end{subfigure}
    \hfill
    \begin{subfigure}[b]{0.47\textwidth}
        \centering
        \includegraphics[width=\textwidth]{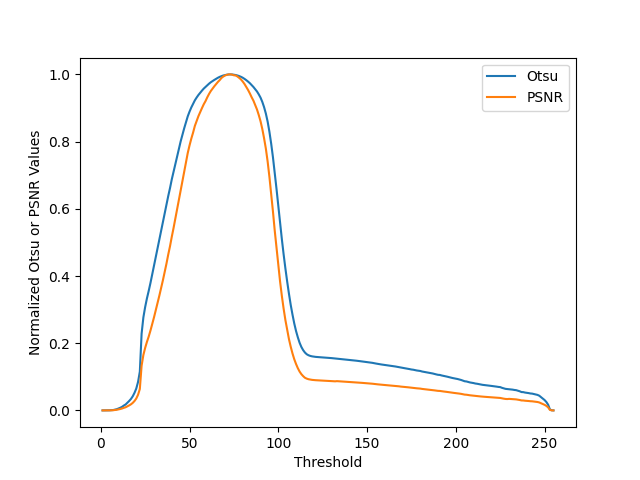}
        \caption{Otsu $\times$ PSNR. Correlation = 0.99.}
    \end{subfigure}


    \begin{subfigure}[b]{0.47\textwidth}
        \centering
        \includegraphics[width=\textwidth]{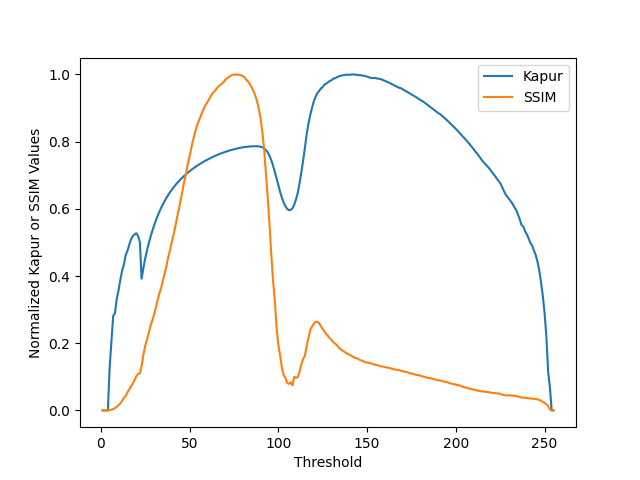}
        \caption{Kapur $\times$ SSIM. Correlation = 0.16.}
    \end{subfigure}
    \hfill
    \begin{subfigure}[b]{0.47\textwidth}
        \centering
        \includegraphics[width=\textwidth]{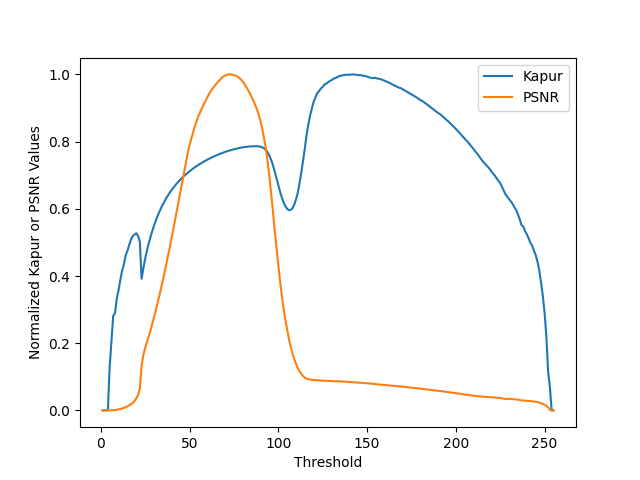}
        \caption{Kapur $\times$ PSNR. Correlation = 0.08.}
    \end{subfigure}
    \caption{Values of Otsu, Kapur, \ac{ssim} and \ac{psnr} on image 37073, from \ac{bsds500}, plotted versus each other. The values of each function/metric are normalized between 0 and 1 for clarity of visualization.}
    \label{fig:indiv_37073}
\end{figure}

\Cref{fig:indiv_28096,fig:indiv_33044,fig:indiv_37073} visualize, for selected images, the relationship between values of the thresholding objective functions (Otsu or Kapur) and the corresponding values of the image quality metrics (\ac{ssim} or \ac{psnr}) evaluated over all possible thresholds. 
These figures illustrate that the degree of correlation depends on how similarly the thresholding objective function and the evaluation metric vary over the threshold space. High correlation values correspond to cases where both quantities increase or decrease in a similar manner, whereas low correlation values indicate divergent trends
An example is \cref{fig:indiv_28096} (d) where the two curves are very similar and their correlation is 0.99, which is very high. An opposite example is \cref{fig:indiv_28096} (e) where the two curves are different/opposite to some sense. In such a case, the correlation coefficient has a low value of 0.14.

For \cref{fig:indiv_28096} (c, d), we can see the curves of Otsu is similar to the curves of both \ac{ssim} and \ac{psnr} and has a high correlation with each of them. However, in \cref{fig:indiv_28096} (e, f), we can see the curves of Kapur is not so similar to the curves of \ac{ssim} and \ac{psnr}. Hence, value of correlation is low. 
For \cref{fig:indiv_33044}, we can see Otsu has low correlation with \ac{ssim} and high correlation with \ac{psnr}, while Kapur has high correlation with \ac{ssim} and low correlation with \ac{psnr}.
Moving to \cref{fig:indiv_37073}, we can see Otsu is highly correlated with both \ac{ssim} and \ac{psnr} but Kapur has a low correlation to \ac{ssim} and \ac{psnr}.

\subsection{Dataset Outputs}
\label{subsec:histogram-cnts}

\begin{figure}
    \centering
    \begin{subfigure}[b]{0.45\textwidth}
        \centering
        \includegraphics[width=\textwidth]{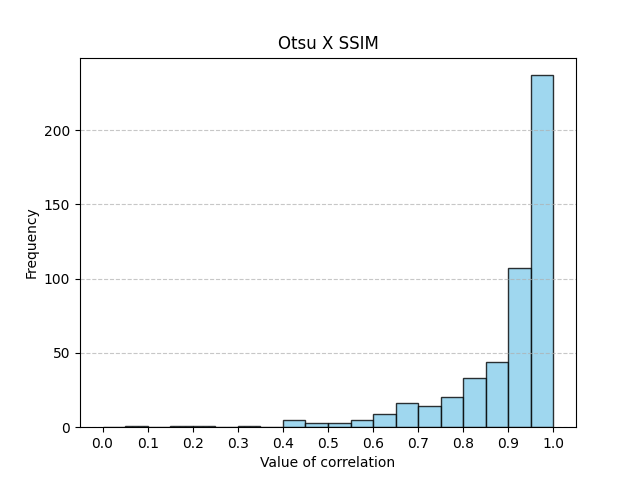}
        \caption{Otsu $\times$ SSIM}
    \end{subfigure}
    \hfill
    \begin{subfigure}[b]{0.45\textwidth}
        \centering
        \includegraphics[width=\textwidth]{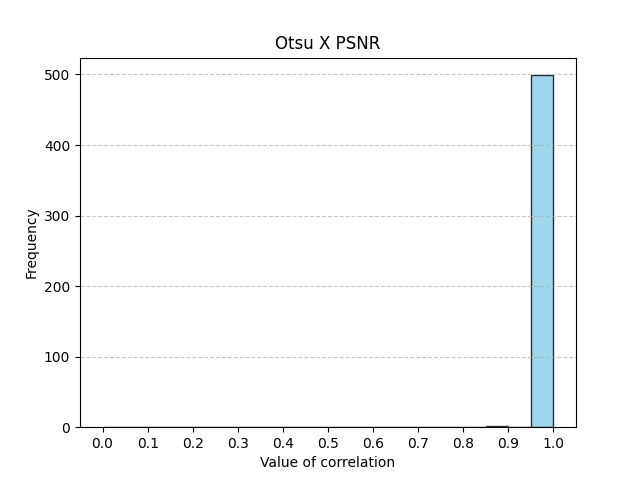}
        \caption{Otsu $\times$ PSNR}
    \end{subfigure}

    \vspace{10pt} 

    \begin{subfigure}[b]{0.45\textwidth}
        \centering
        \includegraphics[width=\textwidth]{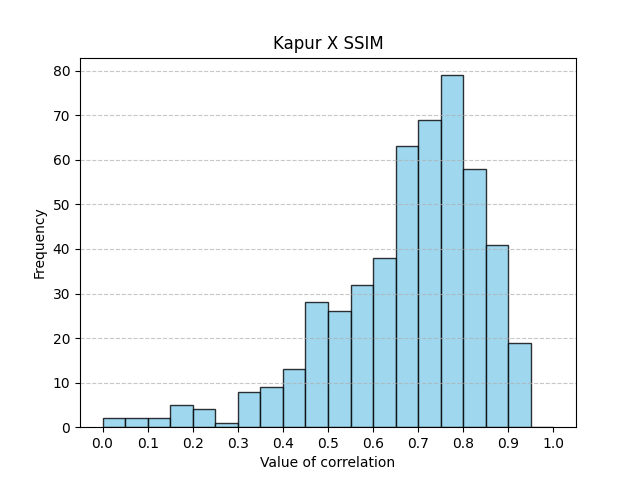}
        \caption{Kapur $\times$ SSIM}
    \end{subfigure}
    \hfill
    \begin{subfigure}[b]{0.45\textwidth}
        \centering
        \includegraphics[width=\textwidth]{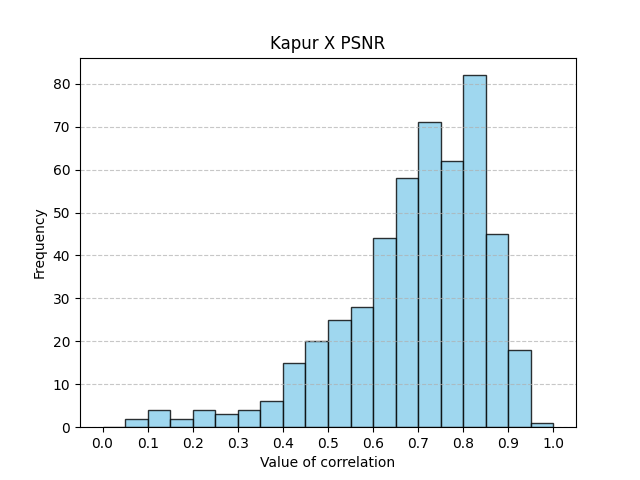}
        \caption{Kapur $\times$ PSNR}
    \end{subfigure}

    \vspace{10pt}

    \caption{Histograms of values of Pearson correlation coefficients over \ac{bsds500} images between each of the thresholding functions of Otsu and Kapur with each of the \ac{ssim} and \ac{psnr} metrics.}
    \label{fig:histogram_comparison}
\end{figure}

\Cref{fig:histogram_comparison} illustrates the distribution of Pearson correlation coefficients computed over the \ac{bsds500} image dataset between the values of Otsu and Kapur functions and the values of \ac{ssim} and \ac{psnr}  metrics. As shown in the figure, the correlation distributions corresponding to Otsu's function are strongly saturated toward the right-hand side, indicating consistently high positive correlation with both \ac{ssim} and \ac{psnr} across most images in the dataset. In contrast, the distributions associated with Kapur's entropy exhibit a more dispersed, approximately normal shape with a mild rightward skew, suggesting weaker and less consistent correlation with the same evaluation metrics. While positive correlations are still frequently observed for Kapur, their magnitudes are generally lower and more variable compared to those obtained with Otsu.

These results indicate that \ac{ssim} and \ac{psnr} are more strongly and consistently aligned with Otsu's thresholding criterion than with Kapur's entropy-based criterion. This systematic difference in correlation behavior suggests the presence of an inherent metric–objective alignment, whereby commonly used quality metrics tend to favor thresholding solutions optimized according to Otsu's function. Consequently, comparative studies that rely solely on \ac{ssim} or \ac{psnr} may implicitly advantage Otsu-based methods over entropy-based ones, regardless of the optimization strategy employed. This observation supports the need for greater caution in interpreting reported performance gains and highlights the importance of considering potential metric-induced bias in the evaluation of multilevel image thresholding methods.

\subsection{Quantitative Results}
\label{subsec:quantit-tables}

For the whole \ac{bsds500} dataset (500 images), statistics about the linear correlation values between each of the objective functions and each of the metrics are summarized in \cref{tab:correlation-results,tab:otsu_vs_kapur_counts}. From \cref{tab:correlation-results}, we can see that Otsu is significantly more correlated to both \ac{ssim} and \ac{psnr} than Kapur. \Cref{tab:otsu_vs_kapur_counts} lists how many images has a higher correlation between Otsu and \ac{ssim} versus how many images has a higher correlation between Kapur and \ac{ssim}. 
It also lists similar data but for the \ac{psnr} metric. We can see from the table that $91\%$ of the \ac{bsds500} dataset images have higher correlation values between Otsu and \ac{ssim} than Kapur and \ac{ssim}. We also see that $100\%$ of the dataset images has a higher correlation for Otsu with \ac{psnr} than Kapur with \ac{psnr}. 
Consequently, for the \ac{bsds500} dataset, \ac{ssim} and \ac{psnr} exhibit a systematic preference in terms of linear correlation toward Otsu function when compared to Kapur.

\renewcommand{\arraystretch}{1.8}
\setlength{\tabcolsep}{8pt}
\begin{table}
\caption{Pearson correlation coefficients ($\text{mean} \pm \text{standard deviation}$) between thresholding objective functions and image quality metrics computed over 500 images from the BSDS500 dataset. 
}
\label{tab:correlation-results}
\centering
\begin{tabular}{c|cc}
\hline
 & {Otsu} & {Kapur} \\
\hline
SSIM   & $\mathbf{0.8959} \pm 0.1312$ & $0.6743 \pm 0.1721$ \\ 
PSNR  & $\mathbf{0.9871} \pm 0.0077$ & $0.6817 \pm 0.1894$ \\
\hline
\end{tabular}
\end{table}

\begin{table}
\caption{Number of images (out of 500) where the correlation of Otsu is higher, or lower than Kapur with SSIM and PSNR metrics.}
\label{tab:otsu_vs_kapur_counts}
\centering
\begin{tabular}{c|cc}
\hline
Metric (M) & Corr(Otsu,M) > Corr(Kapur,M)  & Corr(Kapur,M) > Corr(Otsu,M) \\
\hline
SSIM & \textbf{457 ($\mathbf{91.4\%}$)} & 43 ($8.6\%$)  \\
PSNR & \textbf{500 ($\mathbf{100\%}$)} & 0 ($0\%$) \\
\hline
\end{tabular}
\end{table}
\renewcommand{\arraystretch}{1}
\setlength{\tabcolsep}{6pt}

\section{Conclusions and Future Work}
This study investigated a largely overlooked assumption in multilevel image thresholding research: that commonly used full-reference image quality metrics, particularly SSIM and PSNR, provide an objective and neutral basis for comparing thresholding objective functions. By decoupling the effects of the optimization algorithm from those of the thresholding objective function and the evaluation metric, we demonstrated that this assumption does not generally hold.

Through a systematic correlation analysis conducted over all possible thresholds on the BSDS500 dataset, we showed that SSIM and PSNR exhibit a strong and consistent linear alignment with Otsu's between-class variance criterion, while their alignment with Kapur's entropy-based criterion is significantly weaker and more variable. Quantitatively, Otsu's function achieved substantially higher mean correlation values with both SSIM and PSNR across the dataset, and in the case of PSNR, Otsu outperformed Kapur in terms of correlation for 100\% of the tested images. These findings provide compelling empirical evidence of a metric–objective-function bias inherent in widely used evaluation practices.

The implications of this result are significant for the interpretation of comparative studies in multilevel thresholding. When an optimization algorithm is applied to an objective function that is already favored by the evaluation metric, the resulting method may appear superior regardless of the intrinsic quality of the selected thresholds. Consequently, reported performance gains attributed to novel or hybrid metaheuristic optimizers may, at least in part, reflect this underlying bias rather than genuine improvements in segmentation quality. This observation helps explain the recurring dominance of Otsu-based approaches in SSIM and PSNR-driven comparisons reported throughout the literature.

From a broader perspective, this work calls for greater methodological caution in the evaluation of thresholding algorithms. The choice of evaluation metric should be carefully considered in relation to the optimized objective function, and conclusions drawn from metric-based comparisons should explicitly acknowledge potential alignment effects. Relying on a single quality metric may lead to incomplete or misleading assessments of algorithmic performance.

Several directions for future work naturally follow from this study. First, the present analysis focused on single-threshold (bi-level thresholding). Extending the investigation to true multilevel thresholding scenarios will be essential to confirm whether similar metric biases persist as the dimensionality of the threshold space increases. The scope of this study was kept to bi-level thresholding due to runtime and dataset size constraints. Second, additional evaluation metrics, including perceptual metrics such as \ac{fsim} should be examined to assess whether they exhibit similar alignment behavior with specific thresholding objective functions. Third, extension to other thresholding functions, such as the \ac{met} criterion \cite{kittler,Hegazy_Gabr_B89_2025} or Masi entropy \cite{Masi_2005,adaptive_masi_dp}, can help exploit more alignment and connections between thresholding functions and evaluation metrics. Fourth, expanding the experimental analysis to other datasets, including medical, infrared, and remote sensing imagery, may help determine the extent to which the observed bias is domain-dependent. Finally, these findings motivate the development of more objective or task-aware evaluation frameworks for image thresholding, potentially combining multiple complementary metrics or incorporating application-specific criteria.

In summary, this work highlights a critical yet underexplored factor influencing comparative studies in multilevel image thresholding. By revealing the inherent preference of SSIM and PSNR toward variance-based thresholding objectives, it provides a clearer interpretive lens for existing results in the literature and lays the groundwork for more fair, transparent, and informative evaluation methodologies in future research.

\renewcommand{\bibsection}{\section*{References}}
\bibliographystyle{splncs04nat}
 \bibliography{mybibliography}

\end{document}